\pdfoutput=1

\documentclass[11pt]{article}

\usepackage[final]{EMNLP2023}

\usepackage{times}
\usepackage{latexsym}
\usepackage{color}
\usepackage{tikz}
\usepackage{CJKutf8}
\usepackage{xcolor}
\usepackage{microtype}
\usepackage{mathtools}
\usepackage{adjustbox}
\usepackage{enumitem}
\usepackage{tablefootnote}
\usepackage{multirow}
\usepackage{bm}
\usepackage{amsmath}
\usepackage{amssymb}
\usepackage{booktabs}
\usepackage{ulem}

\usepackage[T1]{fontenc}

\usepackage[utf8]{inputenc}

\usepackage{microtype}

\usepackage{inconsolata}

\newcommand\blfootnote[1]{%
  \begingroup
  \renewcommand\thefootnote{}\footnote{#1}%
  \addtocounter{footnote}{-1}%
  \endgroup
}

\title{Mining Word Boundaries in Speech as Naturally Annotated Word Segmentation Data }

\author{Lei Zhang$^1$, Zhenghua Li$^1$, Shilin Zhou$^1$, Chen Gong$^{1*}$, \\ {\bf Zhefeng Wang$^2$}, {\bf Baoxing Huai$^2$}, {\bf Min Zhang$^1$} \\
$^1$Institute of Artificial Intelligence, School of Computer Science and Technology,\\
Soochow University, Suzhou, China; ~~~$^2$Huawei Cloud, China \\
\texttt{$^1$leizhang.nlp@gmail.com}, \texttt{$^1$slzhou.cs@outlook.com} \\
\texttt{$^1$\{gongchen18, zhli13, minzhang\}@suda.edu.cn} \\
\texttt{$^2$\{wangzhefeng, huaibaoxing\}@huawei.com}
}

\begin{document}
\begin{CJK}{UTF8}{gkai}
\maketitle

\begin{abstract}

Inspired by early research on exploring naturally annotated data for Chinese word segmentation (CWS), and also by recent research on integration of speech and text processing, this work for the first time proposes to mine word boundaries from parallel speech/text data. 
First we collect parallel speech/text data from two Internet sources that are related with CWS data used in our experiments. 
Then, we obtain character-level alignments and design simple heuristic rules for determining word boundaries according to pause duration between adjacent characters. 
Finally, we present an effective  complete-then-train strategy that can better utilize extra naturally annotated data for model training. 
Experiments demonstrate our approach can significantly boost CWS performance in both cross-domain and low-resource scenarios.  
\end{abstract}
\section{Introduction}
\begin{figure}[tb]
\centering
\includegraphics[width=7.5cm]{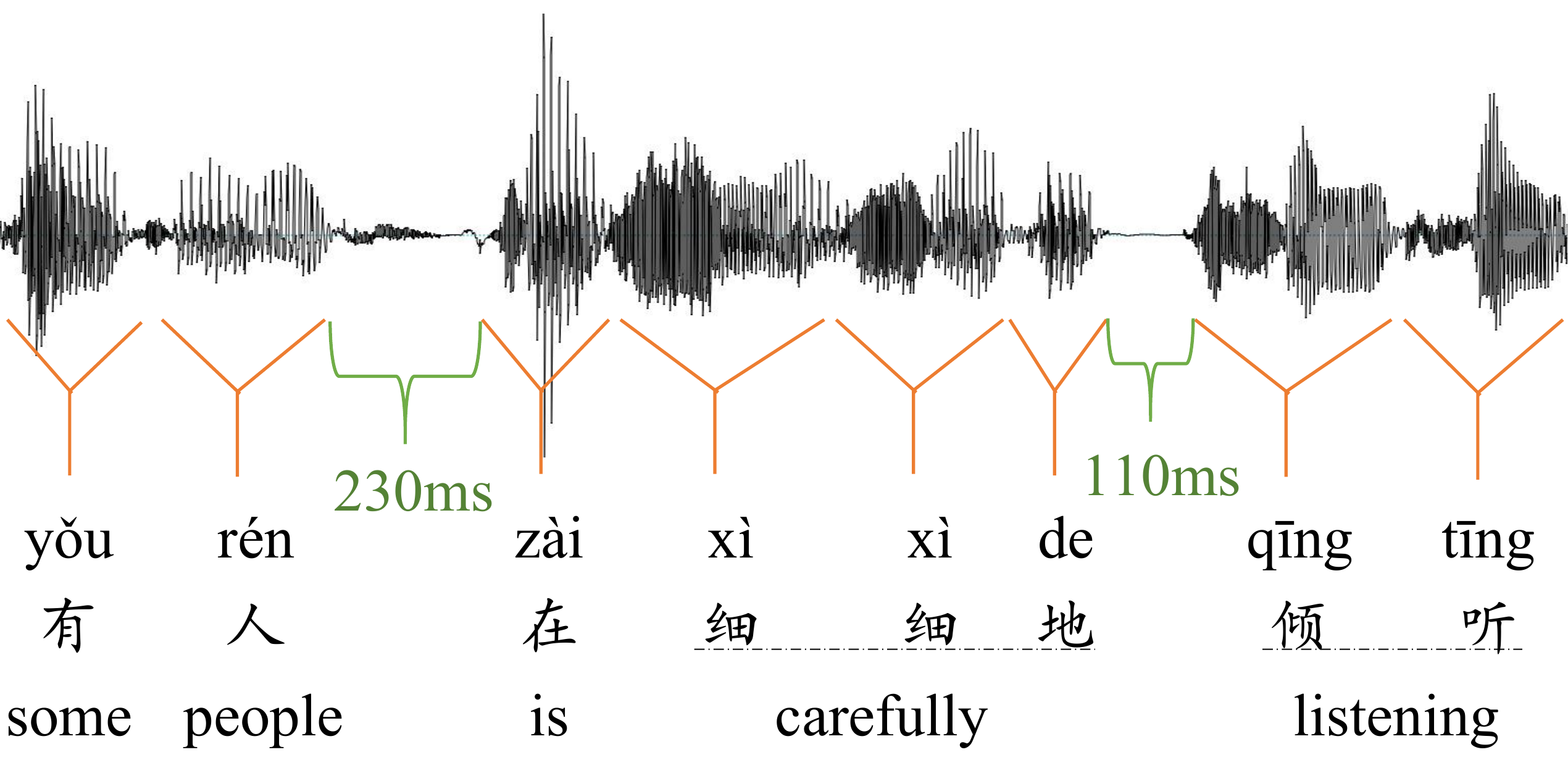}
\caption{An example of speech/text alignment data. The correct segmentation result is  ``有/人/在/细细/地/倾听'', translated as ``some people is carefully listening''. The average duration of all characters is 240 ms in the corresponding speech. 
}
\label{img:example_extract_pause}
\end{figure}

As a fundamental task in Chinese language processing, Chinese word segmentation aims to segment an input character sequence into a word sequence, since words, instead of characters, are the basic meaning unit in Chinese.  
Figure \ref{img:example_extract_pause} gives an example. \blfootnote{$^*$ Corresponding author.}

With the rapid progress of deep learning techniques, especially the proposal of pre-trained language models like BERT \cite{devlin-etal-2019-bert}, CWS models have achieve very high performance when there is abundant training data from the same domain with the test data \citep{tian2020improving,huang2020towards}. Therefore, recent studies on CWS pay more attention to the cross-domain and low-resource scenarios \citep{huang2020joint,ke2021pre}.

Meanwhile, 
considering the high cost of manually annotating high-quality CWS data, it had been an attractive research direction to explore naturally annotated CWS data from different channels. 
For instance, anchor texts in HMLT-format web documents imply reliable word boundaries \citep{jiang2013discriminative,yang2014semi}; domain-aware dictionaries can match words accurately in target domain texts \cite{liu2014domain}. 
These works show that such information can be used as partial annotations for training CWS models. 

Another interesting research line in recent years is the multi-modal integration of speech and texts, mainly due to the adoption of unified model architectures in both speech processing \citep{baevski2020wav2vec,hsu2021hubert} and NLP fields \citep{devlin-etal-2019-bert,lewis2020bart} in the deep learning era. 
These approaches can be broadly divided into three categories, i.e., 
1) using speech as extra features for NLP \citep{zhang2021more}, 2) multi-task learning (MTL) with cross-attention interaction \citep{sui2021large}, and 3) end-to-end language analysis from speech \cite{chen2022aishell}.
Among these, the work of \citet{zhang2021more} is closely related with ours. They extract extra features from speech to enhance CWS on corresponding texts. 

Inspired by both directions of works, we for the first time propose to mine word boundaries from speech as naturally annotated data for CWS. The motivation is that when speaking Chinese, people often pause after finishing some complete meaning in the middle of sentences, to breath or to make the speech easier to understand. Considering that words are the basic meaning unit, we hypothesize that the pause information can be utilized to help CWS. We conduct experiments on two widely used CWS datasets, i.e., Chinese Penn Treebank (CTB) mainly containing newswire texts \cite{xue2005penn}, and the ZX data, which comes from a Chinese web fiction titled as ``ZhuXian'' (``Jade Dynasty'' in English) \cite{zhang2014type}. Experiments show that our approach is useful on both cross-domain and low-resource scenarios. More specifically, our work comprises of the following important components. 

\begin{itemize}[leftmargin=*]
    \item We first collect two parallel speech/text datasets from two sources that are close to to our CWS datasets respectively, followed by careful data cleaning. 
    \item Based on preliminary experiments, we select a conventional GMM-HMM alignment model, which produce more accurate alignments than end-to-end neural network ASR models. We continue training a released model on our data to produce character-level alignments.  
    \item We design two heuristic rules to determine reliable boundaries according to pause duration between adjacent characters. 
    \item We present a simple complete-then-train method that effectively train our model on both labeled CWS data and newly obtained naturally annotated data, after observing that directly employing the approach of \citet{tsuboi2008training} leads to inferior results.  
\end{itemize}

\section{Mining Word Boundaries in Speech}

This section describes how we collect word boundaries from parallel text/speech data, which consists of three steps. 
First, we collect parallel text/speech data from two sources, both of which are closely related with the word segmentation data used in our experiments. 
Second, we utilize a GMM-HMM based model to obtain character-level text/speech alignment. 
Finally, we design a simple strategy to determine word boundaries according to pause duration between characters.

\subsection{Collecting Speech/Text Parallel Data}
\label{section_collecting_data}

For the {newswire} domain (CTB), we follow \citet{zhang2021more} and collect data from the XueXi platform\footnote{https://www.xuexi.cn/}. The platform is founded and maintained by the government and provide very rich multi-media resources on a vast range of channels like politics, education, art, etc. From the topic, we collect formal news and policy articles with both texts and voice recordings, which are mostly close to the newswire domain. We find that some audios are automatically generated by text-to-speech (TTS) models, and filter them based on heuristic rules.  
Finally, we collect 513 high-quality articles after data cleaning, amounting to 178 hours of data. 

As for the cross-domain scenario (ZX), since previous researchers \cite{liu2012unsupervised,zhang2014type} only annotated CWS results for a small fraction of the ZhuXian fiction, we choose to mine word boundaries from the remaining un-annotated texts and use them as naturally annotated CWS data. 
We find several versions of audio books for ZhuXian on the Internet and select one that is of high quality and without background music. 
Finally, we collect 246 chapters after data cleaning and removing those appear in the annotated ZX data, amounting to 144 hours. 

\textbf{Data cleaning.} We apply several data cleaning or filtering strategies to improve data quality. (1) Numbers like ``1200'' are transformed into their Chinese character form like ``一千两百'' (one thousand and two hundred). (2) Silent and special symbols in the texts like punctuation marks are removed.
(3) Irrelevant blanks or noises in the beginning or end of the audio are removed. (4) Audios with background music are discarded. Most importantly, we perform speech/text alignment using MFA (see Section \ref{sec:speech_text_alignment}), and discard the articles or chapters that get poor alignments. 
All audios are processed to be at a sampling frequency of 16kHz.  

\subsection{Character-level Speech/Text Alignment}
\label{sec:speech_text_alignment}

According to the rhythm of Chinese speech, there are usually pauses between adjacent characters and long pauses often occur when some complete meaning is completed. 
We observe that it is clear that the inter-word pause duration is in average longer than intra-word (or at word boundary positions), since words are the basic meaning unit. 
The key idea of this work is to make use of 
pause duration between adjacent characters for mining potential word boundaries. 
The main challenge for this idea is how to obtain accurate character-level alignments between speech signals and the corresponding sentence. 

In the past decade, end-to-end Transformer based models have become the dominate approach due to its superior ASR performance \cite{Gulati2020,Zhang2023GoogleUS,pratap2023scaling}. 
With an extra connectionist temporal classification (CTC) component, the model can explicitly produce alignment. However, our preliminary experiments reveal that the Transformer-CTC based models suffer from a severe peak alignment issue, that means every character is usually aligned to a single speech frame, leaving most of the frames aligned to blanks. 
This finding is consistent with previous results \cite{senior2015acoustic,zeyer2021does}. 

Instead, we employ the Montreal Forced Aligner tookit (MFA, \citealt{mcauliffe2017montreal}) with its GMM-HMM implementation to obtain alignment information between text and speech. 
We employ both monophone and triphone GMMs. For the sake of simplicity, we limit our discussion to monophone GMMs. 

Given a speech, we use the default frame window length of 25ms. For the frame offset, we use 5ms instead of the default 10ms to obtain more fine-grained alignment. For each frame, we extract standard Mel-frequency cepstral coefficients (MFCCs) as the acoustic features. We map Chinese characters to monophones based on a dictionary. Formally, we represent speech as $\mathbf{x}=x_1...x_i...x_n$, where $x_i$ is an MFCC feature vector, and the corresponding transcription as $\mathbf{y}=y_1...y_i...y_{m}$, where $y_i$ denotes a monophone. The objective of GMM-HMM is to determine which frames (e.g., $x_k...x_l$) correspond to a monophone, thus providing the time range for each character. The model works under the unsupervised scenario, and apply the expectation-maximization (EM) algorithm \cite{moon1996expectation} on the training speech/text pairs.

Since the speech/text data we have are at document level, we employ a two-step alignment strategy, which we find can produce more accurate character-level speech/text alignments. First, we continue training a pre-trained model\footnote{https://mfa-models.readthedocs.io/en/latest/acoustic} in  MFA using our document-level data.
Based on the alignments, we segment speech into sentence-level. 

Then, we continue training the model using the sentence-level data. We discard sentences with high-uncertainty alignments.

Finally, we obtain 18,757 sentences from XueXi and 23,568 sentences from ZhuXian. For a sentence comprising $m$ characters, i.e., $\mathbf{s} = c_1...c_i... c_n$, we say $c_i$ is aligned to $x_{b_i}...x_{e_i}$, denoted as $(b_i, e_i)$ only using the beginning and end indices of frames. Then, we can calculate the pause duration between two adjacent characters, for instance $c_i$ and $c_{i+1}$ as follows.
\begin{equation}
d(c_i,c_{i+1}) = (b_{i+1} - e_i) \times \textit{OFS} 
\label{formula:char_interval}
\end{equation}
We can also obtain the duration of pronouncing $c_i$. 
\begin{equation}
d(c_i) = (e_{1} - b_i) \times \textit{OFS} 
\label{formula:char_pronouncing}
\end{equation}
In both equations, $\textit{OFS}=5ms$ is the frame offset.
Figure \ref{img:example_extract_pause} gives an example. 

\begin{figure}[t!]
\centering
\includegraphics[width=7.5cm]{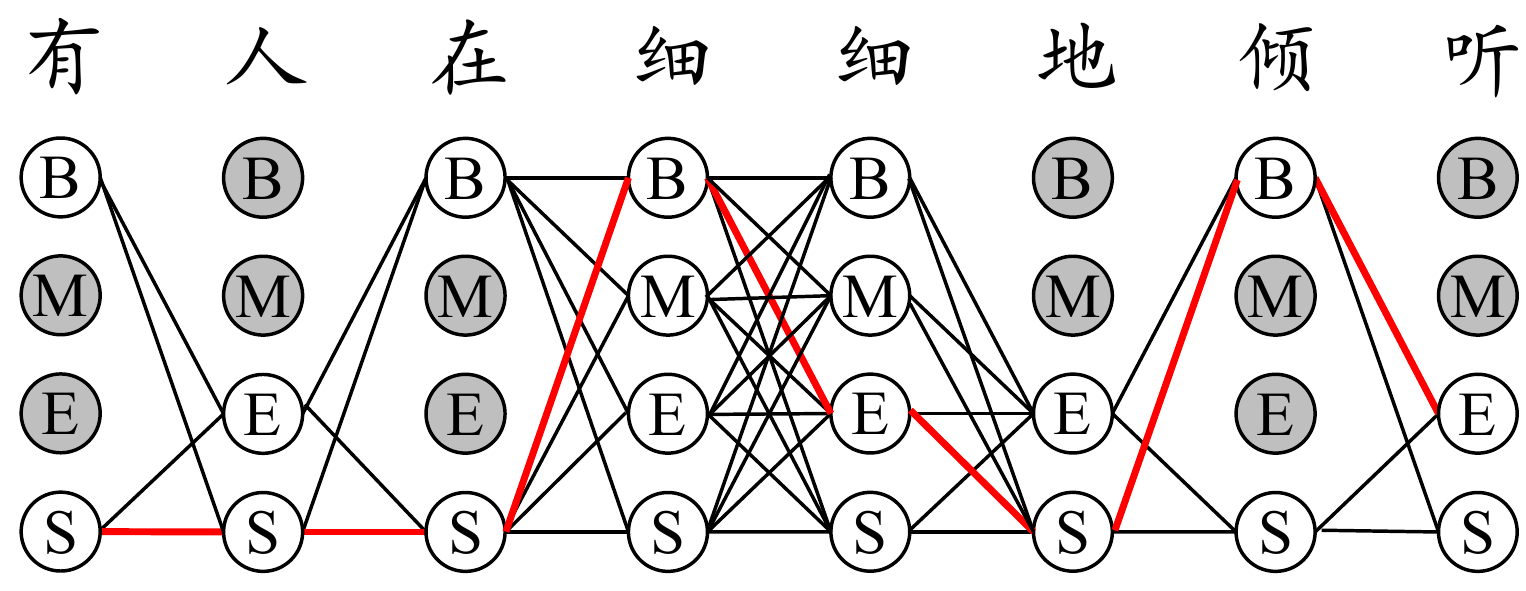}
\caption{Constrained label space for the sentence in Figure 1, in which we obtain two boundaries ``有人/在细细地/倾听''. 
We use the ``BMES'' label scheme, meaning ``begining'', ``middle'', ``end'', and ``single-char'', respectively. 
Illegal labels are marked as gray. The red thick lines present a legal path that may be selected by a model. 
}
\label{img:natural_annotation_constraint}
\end{figure}

\subsection{Word Boundary Extraction}
\label{sec:word_boundart_extraction} 
Our basic intuition is that the longer the pause duration between two adjacent characters is, the more likely there is a word boundary between them, i.e., the former character being the end of a word whereas the latter being the beginning of a word. 
Moreover, considering that speaking speed may be influenced by factors like speakers and contexts, we also take into account the average duration of pronouncing all characters in a sentence. 
After some experimental trials, we decide that there is a word boundary between two characters $c_i$ and $c_{i+1}$ if both of the following two conditions are met. 
\begin{equation}
\begin{split}
\textit{Cond 1:} ~~~ & d(c_i, c_{i+1}) \geq \texttt{MIN} \\ 
\textit{Cond 2:} ~~~ & d(c_i, c_{i+1}) \geq \alpha \times \sum_{j=1}^{n}{ \frac{d(c_j)}{n} }  \\
\end{split}
 \label{formula:vlid_boundary_threshold}
\end{equation}
where $\texttt{MIN}$ is a fixed value as the minimum duration to be a word boundary, and $0 \leq \alpha $ is the minimum ratio against the average pronunciation duration in a sentence. 
Based on our observation, we set them to 50ms and $30\%$ respectively to deal with data from both domains (section \ref{implementation_details} gives more descriptions on the selection of $\texttt{MIN}$ and $\alpha$). As shown in Figure \ref{img:example_extract_pause}, the average duration of pronouncing characters in the sentence is 240ms. 
Therefore, we determine there is a word boundary between two characters when the pause duration between them is longer than $240ms \times 30\% = 72ms$ (both of the above two conditions can be met).

\section{Utilizing Word Boundaries}

\label{section_comeplete_then_train}
\begin{figure}[t!]
\centering
\includegraphics[width=7.5cm]{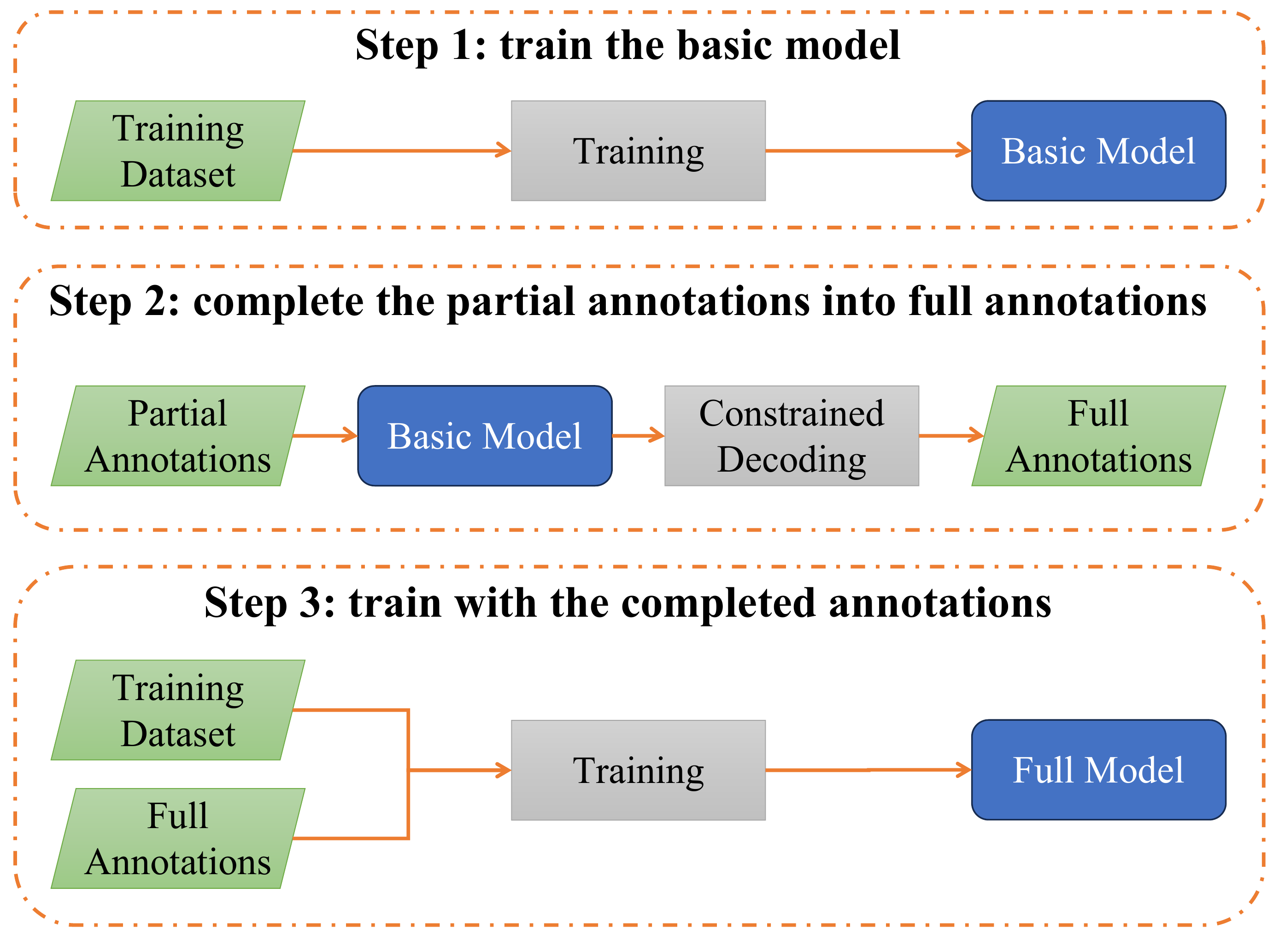}
\caption{The training process of the complete-then-train method. }
\label{img_complete_then_train}
\end{figure}

This section describes how to enhance our basic CWS model with word boundaries in the text data.  

\subsection{As Naturally Annotated CWS Data}
\label{utilizing_naturally_annotated_data}

In fact, quite a few previous studies try to explore word boundaries as naturally annotated CWS data from different channels \cite{jiang2013discriminative,liu2014domain,yang2014semi}. 
Under a sequence labeling framework, word boundaries can be naturally treated as partial annotations and used to construct a constrained label space (aka. lattice). 

Figure \ref{img:natural_annotation_constraint} gives an example. Due to the boundary ``人 (people)/在 (is)'', the left-side char can only be either a single-char word or the end of a word, where as the right-side char can only be either a single-char word or starting a word. 
A similar explanation goes to the second boundary. 
Please notice that it is convenient to constrain labels for the first and last char in the sentence in the same spirit. 

\subsection{The complete-then-train Strategy}
\begin{table}[t!]
    \begin{small}
    \centering
    \setlength{\tabcolsep}{3.5pt}
    \renewcommand{\arraystretch}{1.1}
    \begin{tabular}{lllrrr}
    \toprule 
    Type & Corpus & Amount & Train & Dev & Test\\
    \midrule
    \multirow{4}*{Manual} & \multirow{2}*{CTB5} & \# Sent & 18,104 & 352 & 348 \\
    ~ & ~ & \# Word & 493,932 & 6,821 & 8,008 \\
    \cline{2-6}
    ~ & \multirow{2}*{ZX} & \# Sent & 2,373 & 788 & 1,394 \\
    ~ & ~ & \# Word & 67,648 & 20,393 & 34,355 \\
    \midrule
    \multirow{2}*{Natural} & \multirow{2}*{\emph{SP}} & \# Sent & 42,325 & - & - \\
    ~ & ~ & \# Pauses & 277,676 & - & - \\
    \bottomrule 
    \end{tabular}
    \caption{Statistics of data used in this paper.}  
    \label{experiment_setting}
    \end{small}
\end{table}

Previous works directly use 
the constrained label space in Figure \ref{img:natural_annotation_constraint} to train CWS models via some extension to conventional training objectives. 
Taking conditional random field (CRF) as an example, rather than maximizing the probability of the single gold-standard label sequence, the extended training objective is to maximize the sum total of probabilities of all legal paths in the constrained space, which can be efficiently computed via a variant Viterbi algorithm.

However, our experiments show that directly training model on our data in this way leads to inferior and unsteady performance, and the model is apt to produce single-char words. We suspect the major reason is that 
all characters in the constrained space can be labeled as ``S'' tags, as shown in Figure \ref{img:natural_annotation_constraint}. 

To address the issue, we present a simple yet effective complete-then-train strategy. 
The basic idea is converting partial annotations into full annotations by letting a basic model select an optimal sequence in the constrained space. 
Figure \ref{img_complete_then_train} illustrates the strategy, consisting of three steps. 
First, we train a basic CWS model on the basic CWS training dataset without using naturally annotated data. 
Second, we employ the basic model to complete partial annotations into full ones. More concretely, the basic model selects an optimal label sequence from the constrained space via constrained Viterbi decoding. 
For example, we suppose the model selects the path marked by red thick lines in Figure \ref{img:natural_annotation_constraint}. 
Finally, we use both basic CWS data and completed data to train the full model.

\section{Experiments}
\label{sec:experiments}

\begin{figure}[t!]
\begin{minipage}[b]{.46\linewidth}
  \centering
  \centerline{\includegraphics[width=4.5cm]{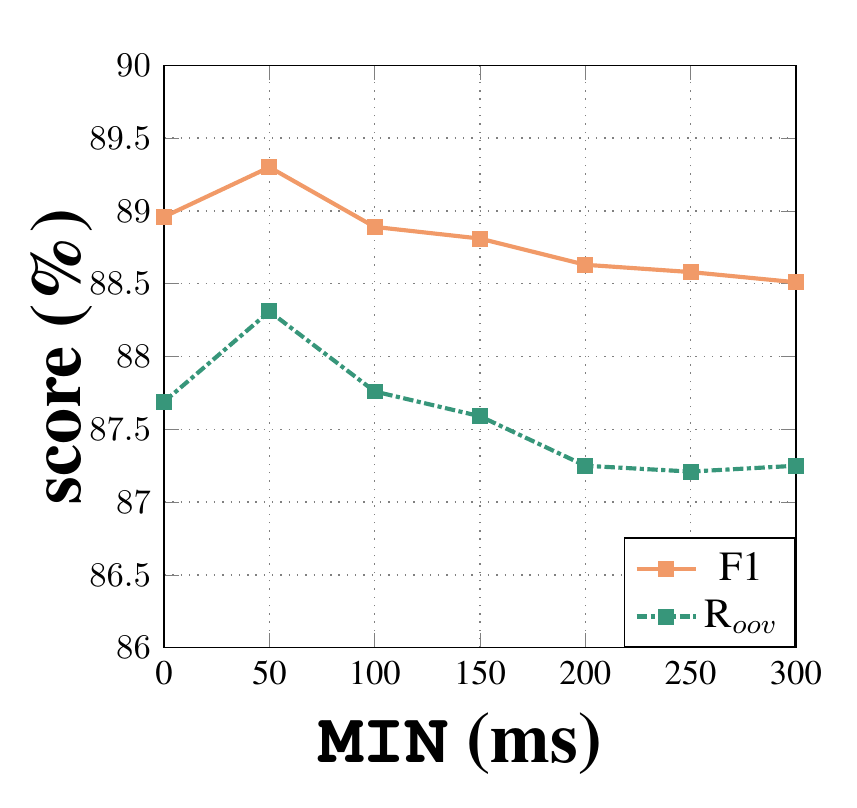}}
\end{minipage}
\hfill
\begin{minipage}[b]{0.46\linewidth}
  \centering
  \centerline{\includegraphics[width=4.5cm]{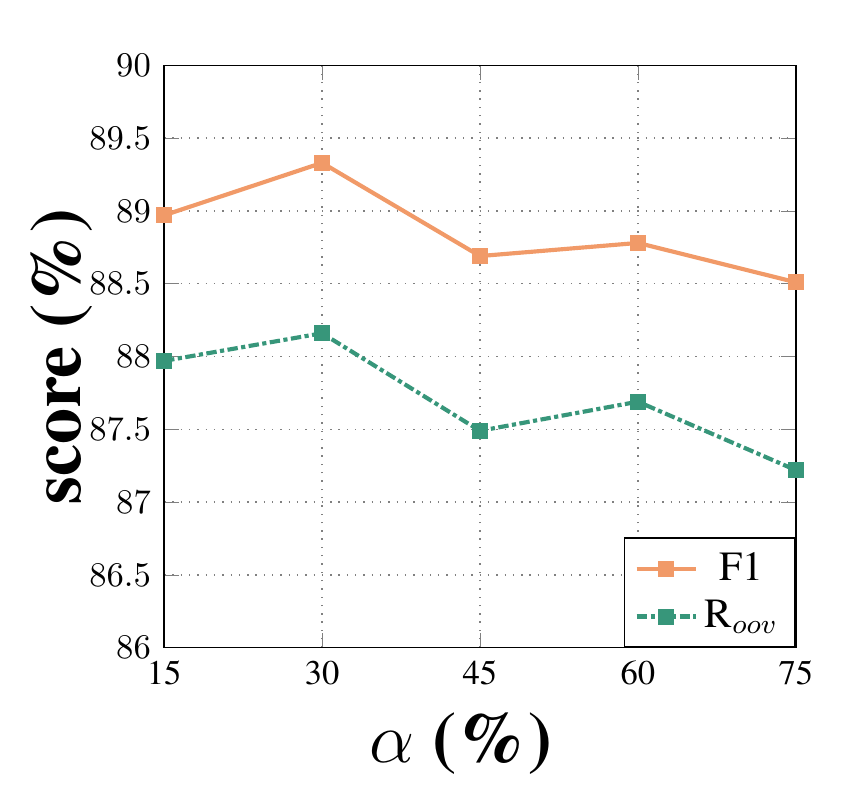}}
\end{minipage}
\caption{Results for the different thresholds $\texttt{MIN}$ and $\alpha$.}
\label{img:threshold}
\end{figure}

\subsection{Implementation Details}
\label{implementation_details}

\paragraph{Data.}

We conduct experiment on both newswire dataset, i.e., CTB5 \cite{xue2005penn}, and web fiction dataset, i.e., ZX \cite{zhang2014type} to evaluate our model in the cross-domain scenario and low-resource scenario. The data statistics are shown in Table \ref{experiment_setting}.

In the cross-domain scenario, we follow previous works on cross-domain CWS \cite{zhang2014type, liu2014domain} to use CTB5-train as the source domain training data, and use ZX-dev and ZX-test as the target domain dev and test data. 

In the low-resource scenario, we evaluate on both CTB5 and ZX datasets with in-domain setting. For training data, we randomly select 1,000 sentences from CTB5-train (denoted as CTB5-1K), and 100 sentences from ZX-train (denoted as ZX-100) as the low-resource training datasets for CTB5 and ZX respectively. For dev/test data, we adopt the original dev/test datasets of CTB5 and ZX respectively.

We use our newly obtained naturally annotated CWS data which is extracted according to speech pause (as illustrated in Section \ref{utilizing_naturally_annotated_data}) as extra training data in both cross-domain and low-resource scenarios, which contains 42,325 sentences and 277,676 word boundaries, denoted as \emph{SP}.

\paragraph{Evaluation metrics.} 

We use the standard precision (P = $\frac{\#Word_{gold \cap pred}}{\#Word_{pred}}$), recall (R = $\frac{\#Word_{gold \cap pred}}{\#Word_{gold}}$), and F1 (F1 = $\frac{2PR}{P+R}$) score for evaluation. In addition, to evaluate the model performance on out-of-vocabulary (OOV) words, we also calculate the recall of OOV words (R$_{oov}$ = $\frac{\#Word^{oov}_{gold \cap pred}}{\#Word^{oov}_{gold}}$).

\paragraph{Model settings.}
We regard CWS as sequence labeling task and implement it with BiLSTM-CRF architecture. Following \citet{wei2021masked}, we impose constraints on candidate paths in both the training and decoding stages.  We use the pre-trained character embedding \emph{glove-6B-100d} released by \citet{pennington2014glove} with a dimension of 100 as the input. We use 3 BiLSTM layers and set the hidden size of LSTM to 300. The mini-batch size is 256 sentences. We use Adam as the optimizer with a learning rate of 0.002. All dropout ratios are 0.1.
Early stopping is triggered when the peak performance on dev data does not increase in 10 epochs. 

To decide the value of parameters $\texttt{MIN}$ and $\alpha$ in Equation \ref{formula:vlid_boundary_threshold}, we first fix $\alpha$ to 0 and compare the performance under different parameters $\texttt{MIN}$ to pick the best $\texttt{MIN}$. Then, we fix the best $\texttt{MIN}$ and compare the results with different $\alpha$ to choose the best $\alpha$. Finally, we set $\texttt{MIN}$ to 50 and set $\alpha$ to 30 respectively according to the results in Figure \ref{img:threshold}.

We run each model for 3 times with different random seeds and report the average results.

\begin{table}
    \begin{small}
    \centering
    \setlength{\tabcolsep}{3.5pt}
    \renewcommand{\arraystretch}{1.1}
    \begin{tabular}{lcccc}
    \toprule 
    Models & P & R & F & R$_{oov}$ \\
    \midrule
    CTB5 & 87.75 & 88.05 & 87.90$_{\pm0.37}$ & 86.59$_{\pm0.54}$ \\
    ~~~~+ \emph{SP} & 89.03 & 89.39 & \textbf{89.21}$_{\pm0.12}$ & \textbf{88.03}$_{\pm0.13}$ \\
    \midrule
    CTB5 + \emph{Lex} ~~~~ & 90.68 & 90.94 & 90.80$_{\pm0.29}$ & 90.17$_{\pm0.09}$ \\
    ~~~~+ \emph{SP} & 91.94 & 92.07 & \textbf{92.01}$_{\pm0.16}$ & \textbf{91.51}$_{\pm0.17}$ \\
    \midrule
    \midrule
    \multicolumn{3}{l}{Zhang et al. (2014) + \emph{Lex}} & 88.34 & - \\ 
    \multicolumn{3}{l}{Liu et al. (2014) + \emph{Lex}} & 90.36 & 80.69 \\
    \multicolumn{3}{l}{~~~~+ \emph{WIKI} (5.5M sent)}  & 90.63 & 84.88 \\
    \bottomrule 
    \end{tabular}
    \caption{Results in the cross-domain scenario.
    }
    \label{cross_domain}
    \end{small}
\end{table}

\subsection{Results in Cross-domain Scenario}
\label{section_cross_domain}

Table \ref{cross_domain} shows the results in cross-domain scenario.

From the first major row, we can see that the model (``+\emph{SP}'') using our natural annotations from speech as extra training data can bring a significant improvement by 1.31 in F1 score and 1.44 in R$_{oov}$ compared with the model (``CTB5'') which is only trained on CTB5. This demonstrates that our approach of mining word boundaries from speech pause is feasible and can provide an effective solution for alleviating data sparseness problem in the cross-domain scenario.

Then, we further augment the training data of the models in the first major row by adding an addition naturally annotated data denoted as ``\emph{Lex}'' released by \citet{liu2014domain} to compared with previous cross-domain CWS works. ``\emph{Lex}'' is obtained by matching the sentences in unlabeled Zhuxian texts with a lexicon, and contains 32,023 sentences. The second major row shows the results.
We can see that when adding ``\emph{Lex}'' in training data, the further utilization of our natural annotations ``\emph{SP}'' can still achieves a large boost of 1.19 in F1 score and 1.34 in R$_{oov}$, showing that the word boundary information mined from speech can complement the word information obtained from lexicon, further verifying the effectiveness of our method.

When comparing with previous works, as shown in the third major row, our model trained with ``\emph{Lex}'' and ``\emph{SP}'' achieves the highest performance.
It is worth mentioning that although \citet{liu2014domain} use a much larger amount dataset \emph{WIKI}, which contains 5.5M sentences, as natural annotations for training, our model still gains better results.
This demonstrates that our natural annotations from speech can bring more guidance to CWS.

\begin{table}
    \begin{small}
    \centering
    \setlength{\tabcolsep}{5.0pt}
    \renewcommand{\arraystretch}{1.1}
    \begin{tabular}{lcccc}
    \toprule
    Data & P & R & F & R$_{oov}$ \\
    \midrule
    CTB5-1K & 87.87 & 89.40 & 88.63$_{\pm0.40}$ & 88.62$_{\pm0.36}$\\
    ~~~+ \emph{SP} & 90.63 & 91.55 & \textbf{91.09}$_{\pm0.02}$ & \textbf{90.95}$_{\pm0.20}$ \\
    \midrule
ZX-100 & 82.07 & 84.06 & 83.05$_{\pm0.19}$ & 83.11$_{\pm0.06}$ \\
    ~~~+ \emph{SP} & 83.96 & 85.61 & \textbf{84.80}$_{\pm0.10}$ & \textbf{84.91}$_{\pm0.51}$ \\
    \midrule
    \midrule
    CTB5 & 96.23 & 97.09 & 96.83$_{\pm0.29}$ & 96.76$_{\pm0.36}$ \\
    ~~~+ \emph{SP} & 96.77 & 97.47 & \textbf{97.12}$_{\pm0.05}$ & \textbf{97.14}$_{\pm0.21}$\\
    \midrule
    ZX & 92.79 & 93.23 & 93.01$_{\pm0.26}$ & 92.87$_{\pm0.40}$\\
    ~~~+ \emph{SP} & 93.37 & 93.70 & \textbf{93.53}$_{\pm0.09}$ & \textbf{93.40}$_{\pm0.15}$ \\
    \bottomrule
    \end{tabular}
    \caption{Results in low-resource scenario. 
    }
    \label{low_resource}
    \end{small}
\end{table} 
\subsection{Results in Low-resource Scenario}

The results in low-resource scenario are shown in the first two major rows of Table \ref{low_resource}. We also present the results in rich-resource scenario in the last two major rows of Table \ref{low_resource} for reference.

In the newswire domain, as shown in the first major row of Table \ref{low_resource}, the model augmented with our naturally annotated data ``+\emph{SP}'' can bring a 2.46 boost in F1 score and 2.33 in R$_{oov}$ compared with the model only trained on the low-resource CTB5-1K.
Even using all the CTB5 training set, using \emph{SP} can still leads to consistent improvements in F1 and R$_{oov}$ respectively.
In the web fiction domain, the further utilization of \emph{SP} improves the F1 score and R$_{oov}$ significantly when training on ZX-100.
Moreover, our natural annotated data on the full ZX dataset can still bring an improvement.

The above analysis shows that when human annotation resources are scarce, our natural annotations obtained from speech can bring great improvements to CWS, effectively alleviating data sparseness problems. 

\section{Analysis}
\label{sec:analysis}
In order to understand the improvements introduced by our method, we conduct detailed analysis from different perspectives, taking the performance in the cross-domain scenario as case study.

\subsection{Ablation Study}
We conduct ablation study to understand the contribution of each individual component in our work. Table \ref{ablation_boost_source} shows the test results. 
\paragraph{Impact of completing with constraint.}
\begin{table}
    \begin{small}
    \centering
    \setlength{\tabcolsep}{1.3pt}
    \renewcommand{\arraystretch}{1.1}
    \begin{tabular}{lcccc}
    \toprule 
    ~ & P & R & F & R$_{oov}$ \\
    \midrule
    CTB5+\emph{SP}&89.03 & 89.39 & \textbf{89.21}$_{\pm0.12}$ & \textbf{88.03}$_{\pm0.13}$ \\
    \hspace{7pt}w/o label constraint&87.13 & 87.52 & 87.33$_{\pm0.28}$ & 85.89$_{\pm0.25}$ \\
    \hspace{7pt}directly-train strategy &87.79 & 88.66 & 88.23$_{\pm0.49}$ & 87.10$_{\pm0.42}$ \\
    CTB5 &87.75 & 88.05 & 87.90$_{\pm0.37}$ & 86.59$_{\pm0.54}$ \\
    \bottomrule 
    \end{tabular}
    \caption{Results of ablation study.}
    \label{ablation_boost_source}
    \end{small}
\end{table}

As illustrated in Section \ref{utilizing_naturally_annotated_data}, we use a basic model to complete the partial annotations in \emph{SP} into full annotations, performing constrained decoding to ensure the completed full annotations not violating existing partial annotations, and then take the completed \emph{SP} as extra training data (row ``CTB5+\emph{SP}''). The ``w/o label constraint'' row presents the result that disregard our mined word boundaries in \emph{SP} and obtain pseudo fully annotated \emph{SP} via a basic model with normal decoding, instead of performing constrained decoding on constrained label space as shown in Figure \ref{img_complete_then_train}. 
We observe that the result of ``w/o constraint''  decreases dramatically, even inferior to the result without using additional training data (row ``CTB5''), probably because the pseudo fully annotated \emph{SP} without constraint contains much noise and can disturb the model training. This verifies that our mined word boundaries in \emph{SP} can effectively provide additionally guidance to CWS model training.

\paragraph{Impact of training strategy.}
To analyze the impact of training strategy, we replace the complete-then-train strategy with the directly-train strategy and keep other model settings unchanged. Directly-train strategy learns from partial annotations directly, based on the idea of maximizing the probabilities of all the possible label sequences that does not violate existing partial annotations, i.e., maximizing the probabilities of all the legal paths in the constrained label space in Figure \ref{utilizing_naturally_annotated_data}. Such ``directly-train strategy'' causes a performance drop of 0.98 and 0.93 in F1 score and R$_{oov}$ respectively. The probably reason is that it is legal for all the characters to be predicted as ``S'' label in the constrained label space, as shown in Figure \ref{utilizing_naturally_annotated_data}, leading the model tends to incorrectly producing many single-char words. The complete-then-train strategy can effectively address the above issue.

\label{sec:exp_data_scale}
\begin{figure}[t!]
\centering
\includegraphics[width=7.7cm]{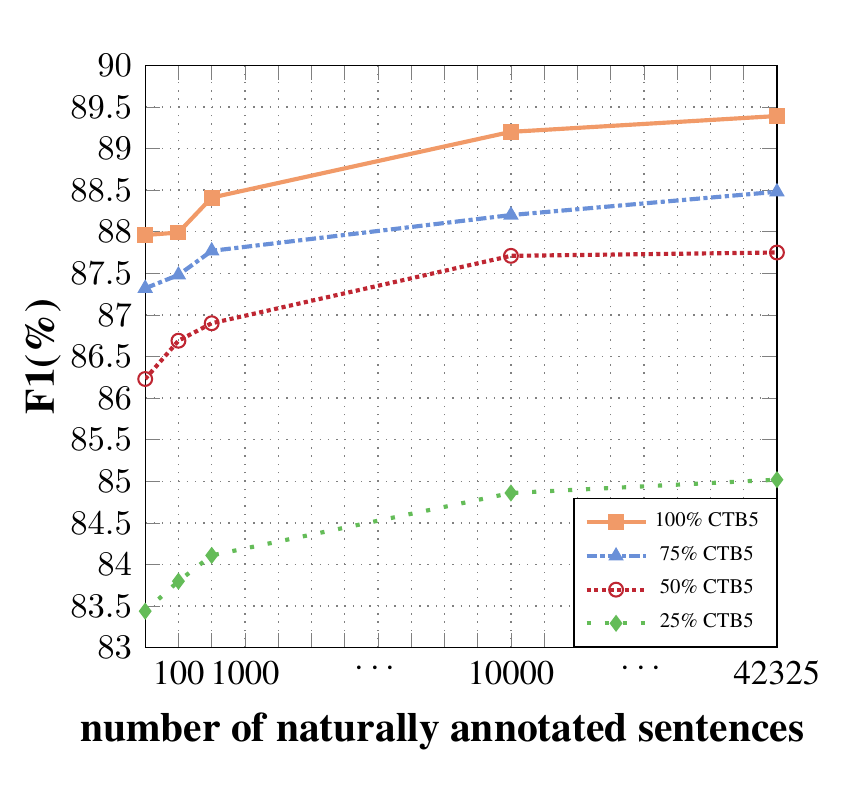}
\caption{Influence of utilizing different amount of additional \emph{SP} data on different CTB5 data scales.}
\label{img:diff_data_scale}
\end{figure}

\subsection{Performance Regarding Data Scale}
We conduct detailed analysis to understand the influence of the amount of our natural annotations from speech. 
Figure \ref{img:diff_data_scale} shows the results.
For the four curves, we use 100\%, 75\%, 50\% and 25\% of the CTB5 training set as the training set. 
In each curve, We randomly sample 100, 1,000, 10,000 and all the sentences from our naturally annotated \emph{SP} dataset and merge them into the CTB5 training set, respectively.

We observe that using more \emph{SP} dataset for training can bring consistently better performance for all four curves, verifying the effectiveness of \emph{SP} in improving CWS models. From another aspect, by comparing the four curves, we can see that the model trained on 25\% CTB5 gains more improvements when leveraging additional \emph{SP} data than the models trained on 50\% CTB5, 75\% CTB5, and 100\% CTB5. This demonstrate that \emph{SP} is able to alleviate data sparseness problem, especially when the size of training data is low-resource.

\subsection{Performance under Different Neural Architectures}

We present the experimental results to investigate the effectiveness of our proposed method when implemented with different model architectures.
For the BERT-based architecture, we utilized the ``BERT-Base-Chinese''\footnote{https://huggingface.co/bert-base-chinese} released by Hugging Face.
Table \ref{ablation_encoder} shows the performance under BiLSTM-, Transformer- and BERT-based architecture. 
We can see that the model encoded with BiLSTM architecture outperforms that with Transformer architecture, and the utilization of \emph{SP} improves the model performance by large margin under both of the two neural architectures. When employing the contextualized character representations obtained from BERT encoder, the performance of both ``CTB5'' and ``+\emph{SP}'' boost. Although utilizing \emph{SP} brings fairly small improvement in F1 score, probably due to the high F1 score already, it brings larger improvement in  R$_{oov}$, verifying the effectiveness of our method in alleviating the OOV problem.

\begin{table}
    \begin{small}
    \centering
    \setlength{\tabcolsep}{4.5pt}
    \renewcommand{\arraystretch}{1.1}
    \begin{tabular}{lccccc}
    \toprule 
    Models & P & R & F & R$_{oov}$ \\
    \midrule
    \textbf{BiLSTM} \\
    CTB5 & 87.75 & 88.05 & 87.90$_{\pm0.37}$ & 86.59$_{\pm0.54}$ \\
    ~~~~+ \emph{SP} & 89.03 & 89.39 & \textbf{89.21}$_{\pm0.12}$ & \textbf{88.03}$_{\pm0.13}$ \\
    \midrule
    \textbf{Transformer} \\
     CTB5 & 83.32 & 84.33 & 83.82$_{\pm0.34}$ & 79.05$_{\pm0.27}$ \\
    ~~~~+ \emph{SP} & 85.18 & 85.63 & \textbf{85.40}$_{\pm0.25}$ & \textbf{81.17}$_{\pm0.38}$ \\
    \midrule
    \textbf{BERT} \\
    CTB5 & 94.03 & 94.31 & 94.17$_{\pm0.06}$ & 93.14$_{\pm0.16}$ \\
    ~~~~+ \emph{SP} & 94.25 & 94.22 & \textbf{94.23}$_{\pm0.06}$ & \textbf{93.48}$_{\pm0.21}$ \\
    \bottomrule 
    \end{tabular}
    \caption{Ablation experiment results of different encoders.
    }
    \label{ablation_encoder}
    \end{small}
\end{table}

\section{Related Works}
\label{sec:related_works}

\subsection{Integrated Speech and Text Processing} 

In the deep learning, the Transformer-based model architecture becomes popular in both speech processing and NLP fields. 
The same architecture makes it convenient to process speech and textual data in an integrated manner. Intuitively, speech and text can provide complementary useful features. We summarize recent works into three categories.  

\paragraph{(1) Speech as extra features for NLP.}
The most straightforward way is to extract features from speech and use them as extra inputs for an NLP model. \citet{zhang2021more} present an interesting pioneer effort and use speech features to help CWS, which is closely with our work. Their approach requires parallel speech/text data in both training and test phases, with WS annotations and the character/frame alignments. They manually annotate 250 sentences and split them into training/test data. Experiments show that extra speech features are beneficial. 

Different from their work, ours emphasis on the use of pause information in speech. We do not need WS annotations for the text data and automatically derive character/frame alignments. In the test phase, our CWS model performs on on text data, rather than parallel speech/text data. 

\paragraph{(2) MTL 
with cross-attention interaction.} 
Given parallel speech/text data, \citet{sui2021large} present a multi-task learning approach that performs NER and ASR at the same time. They first use separate encoders for the two types of inputs, and then employ the cross-attention mechanism to achieve multi-model interaction. 

\paragraph{(3) End-to-End language analysis from speech. }
Several works propose to directly derive language analysis results from speech inputs in an end-to-end manner. 
\citet{ghannay2018end} embed NE labels into texts and train a model that transcribes speech into texts and treats NE labels as normal tokens. They conduct experiments on French NER. 
\citet{yadav2020end} apply the approach to English NER and propose a new label embedding scheme. 
\citet{chen2022aishell} present a Chinese datasets of parallel speech/text data with NE annotations, and 
systematically compare the pipeline and end-to-end approaches.

\citet{wu2022towards} propose an end-to-end relation extraction model that transcribes speech into (entity, entity, relation) triples, and  totally ignores the full text (not performing ASR). However, their experiments show that the end-to-end approach is inferior to the pipeline model, i.e., first ASR and then relation extraction on texts.

\paragraph{Utilizing speech pauses.} 
\citet{fleck2008lexicalized} make use of speech pauses to help English ASR, and more specifically to help transforming phonemes into words. The pauses are output by a previous ASR component and are embedded in the phoneme sequence. They propose to use the pauses to segment the phoneme sequence into several fragments and transform them into words separately. 

\subsection{Naturally annotated CWS data}

\paragraph{Mining naturally annotated data.}
Previous studies try to mine naturally annotated CWS data from different channels. 
\citet{jiang2013discriminative} hypothesize that anchor texts (i.e., for hyperlinks) in HTML-format web documents are very likely to correspond to complete meaning units, and thus can be explored to obtain at least two word boundaries. 
In the cross-domain scenario, \citet{liu2014domain} use a domain-related dictionary and perform maximum matching on unlabeled target-domain text, treating matched texts as annotated words. 

\paragraph{Utilizing naturally annotated data.} 
Above naturally annotated data are in two forms. In the first form, some word boundaries in the sentence are given, whereas in the second, some words are given. 
Both forms can be treated as partial annotations, in contrast to full annotations, and be encoded as constrained label space as shown in Figure xx.

\citet{jiang2013discriminative} propose a constrained decoding approach to learn from partially annotated data with word boundaries. 
They use a max-margin training loss. For each training sentence, they first obtain an optimal label sequence from the constrained space and use it as gold-standard reference in an online fashion. 

\citet{liu2014domain} and \citet{yang2014semi} employ the CRF loss and follow
\citet{tsuboi2008training} to extend the loss for learning from partial/incomplete annotations.
In this work, we also use this approach, but obtain inferior performance probably due to the issue of pervasive ``S'' labels. 
We propose a simple yet effective complete-then-train strategy. 

\section{Conclusion}
\label{sec:conclusion}

This paper for the first time proposes to mine word boundaries from speech/text data as extra naturally annotated training data for CWS. First, we collect large-scale speech/text data from both newswire and web fiction domains. 
Then, we perform character-level alignment on the speech/text data, and determine word boundaries in the data according to the pause duration between two adjacent characters.
Finally, we employ a complete-then-train method to leverage our natural annotations as extra training data to improve CWS performance in cross-domain and low-resource scenarios. Extensive experiments and detailed analysis show that our proposed approach of mining word boundaries in speech as naturally annotated data is effective in improving CWS performance in both cross-domain and low-resource scenarios.

\section*{Limitations}

Our presented work may be further improved in the aspect of two possible limitations. First, our approach heavily relies on accurate alignment between speech and texts. We find that the end-to-end Transformer-CTC model is incapable of producing accurate alignments possibly because it mainly focuses on the ASR task and  only implicitly models the alignment task. After several trails, we finally employ the conventional GMM-HMM model in the MFA toolkit. Although the alignments are satisfactory, we believe that more potential of our approach can be exploited if we can optimize the alignment model and have more accurate alignments.  

Second, we find that our approach is effective when we can obtain parallel speech/text data that is relatively clean and shares similar genres with the test data. However, collecting such qualified data is not easy and requires a lot of dirty work. For instance, we find that the widely used AISHELL-1 data is not useful. Besides the two factors mentioned just now, another possible reason is that it contains a large portion of disfluent sentences. 
\bibliography{anthology,custom}
\bibliographystyle{acl_natbib}

\end{CJK}
\end{document}